\documentclass[conference]{IEEEtran}
\IEEEoverridecommandlockouts
\usepackage{graphicx} 
\usepackage{pifont}
\usepackage{amsmath}
\usepackage{amssymb}  
\usepackage{amsfonts}
\usepackage{algorithmic}
\usepackage{array}
\usepackage{multicol}

\usepackage{textcomp}
\usepackage{stfloats}
\usepackage{url}
\usepackage{verbatim}
\usepackage{graphicx}
\usepackage{caption}
\usepackage{subcaption}
\usepackage[ruled, vlined, linesnumbered]{algorithm2e}
\usepackage{cite}
\usepackage{booktabs}
\usepackage{amssymb,amsfonts,bm}
\usepackage{amsmath,tikz}
\usepackage{color}
\usepackage{mathtools}
\usepackage[hidelinks]{hyperref} 
\usepackage{rotating}
\usepackage{blkarray}
\usepackage{physics}
\usetikzlibrary{arrows}
\usepackage{makecell}

\usepackage{amsthm}
\usepackage{comment}
\usepackage{multirow}
\usepackage{geometry}
\graphicspath{{images/}}

\usepackage{caption}
\makeatletter
\newcommand{\frameglobal}{\mathcal{G}}
\newcommand{\framebody}{\mathcal{B}}
\newcommand{\frameproj}{{\mathcal{B}_{\Pi}}}
\def\frG{\frameglobal}
\def\frB{\framebody}
\def\frP{\frameproj}
\newcommand{\globX}{x_\frG}
\newcommand{\globY}{y_\frG}
\newcommand{\globZ}{z_\frG}
\newcommand{\globroll}{{\phi}_\frG}
\newcommand{\globpitch}{{\theta}_\frG}
\newcommand{\globyaw}{{\psi}_\frG}

\newcommand{\globXvec}{\vec{x}_\frG}
\newcommand{\globYvec}{\vec{y}_\frG}
\newcommand{\globZvec}{\vec{z}_\frG}

\newcommand{\globVX}{\dot{x}_\frG}
\newcommand{\globVY}{\dot{y}_\frG}

\newcommand{\globAX}{\ddot{x}_\frG}
\newcommand{\globAY}{\ddot{y}_\frG}

\newcommand{\globW}{\omega_\frG}
\newcommand{\globWd}{\dot{\omega}_\frG}

\newcommand{\vehicleXvec}{\vec{x}_\frB}
\newcommand{\vehicleYvec}{\vec{y}_\frB}
\newcommand{\vehicleZvec}{\vec{z}_\frB}

\newcommand{\projXvec}{\vec{x}_{\frameproj}}
\newcommand{\projYvec}{\vec{y}_{\frameproj}}

\newcommand{\vehicleVX}{\dot{x}_\frB}
\newcommand{\vehicleAX}{\ddot{x}_\frB}
\newcommand{\vehicleAY}{\ddot{y}_\frB}
\newcommand{\vehicleW}{\omega_\frB}
\newcommand{\vehicleWd}{\dot{\omega}_\frB}

\newcommand{\projVX}{\dot{x}_{\frameproj}}

\newcommand{\projW}{\omega_{\frameproj}}

\begin{document}


  
\title{\LARGE \bf
Energy-Constrained Navigation for Planetary Rovers under Hybrid RTG-Solar Power\\

\author{Tianxin Hu$^{*}$, 
Weixiang Guo$^{*}$, 
Ruimeng Liu, 
Xinhang Xu, 
Rui Qian, 
Jinyu Chen, 
Shenghai Yuan$^{\dagger}$, 
and Lihua Xie%
\thanks{$^{*}$ Equal Contribution.}%
\thanks{$^{\dagger}$ Corresponding Author.}%
\thanks{All authors are with the Centre for Advanced Robotics Technology Innovation (CARTIN), School of Electrical and Electronic Engineering, Nanyang Technological University, 50 Nanyang Avenue, Singapore 639798.}%
\thanks{Emails: \{shyuan, elhxie\}@ntu.edu.sg}%
}


}

\maketitle

\begin{abstract}
Future planetary exploration rovers must operate for extended durations on hybrid power inputs that combine steady radioisotope thermoelectric generator (RTG) output with variable solar photovoltaic (PV) availability. While energy-aware planning has been studied for aerial and underwater robots under battery limits, few works for ground rovers explicitly model power flow or enforce instantaneous power constraints. Classical terrain-aware planners emphasize slope or traversability, and trajectory optimization methods typically focus on geometric smoothness and dynamic feasibility, neglecting energy feasibility.
We present an energy-constrained trajectory planning framework that explicitly integrates physics-based models of translational, rotational, and resistive power with baseline subsystem loads, under hybrid RTG–solar input. By incorporating both cumulative energy budgets and instantaneous power constraints into SE(2)-based polynomial trajectory optimization, the method ensures trajectories that are simultaneously smooth, dynamically feasible, and power-compliant.
Simulation results on lunar-like terrain show that our planner generates trajectories with peak power within 0.55\% of the prescribed limit, while existing methods exceed limits by over 17\%. This demonstrates a principled and practical approach to energy-aware autonomy for long-duration planetary missions.
\end{abstract}


\section{Introduction}
\begin{figure}
    \centering
    \includegraphics[width=1.0\linewidth]{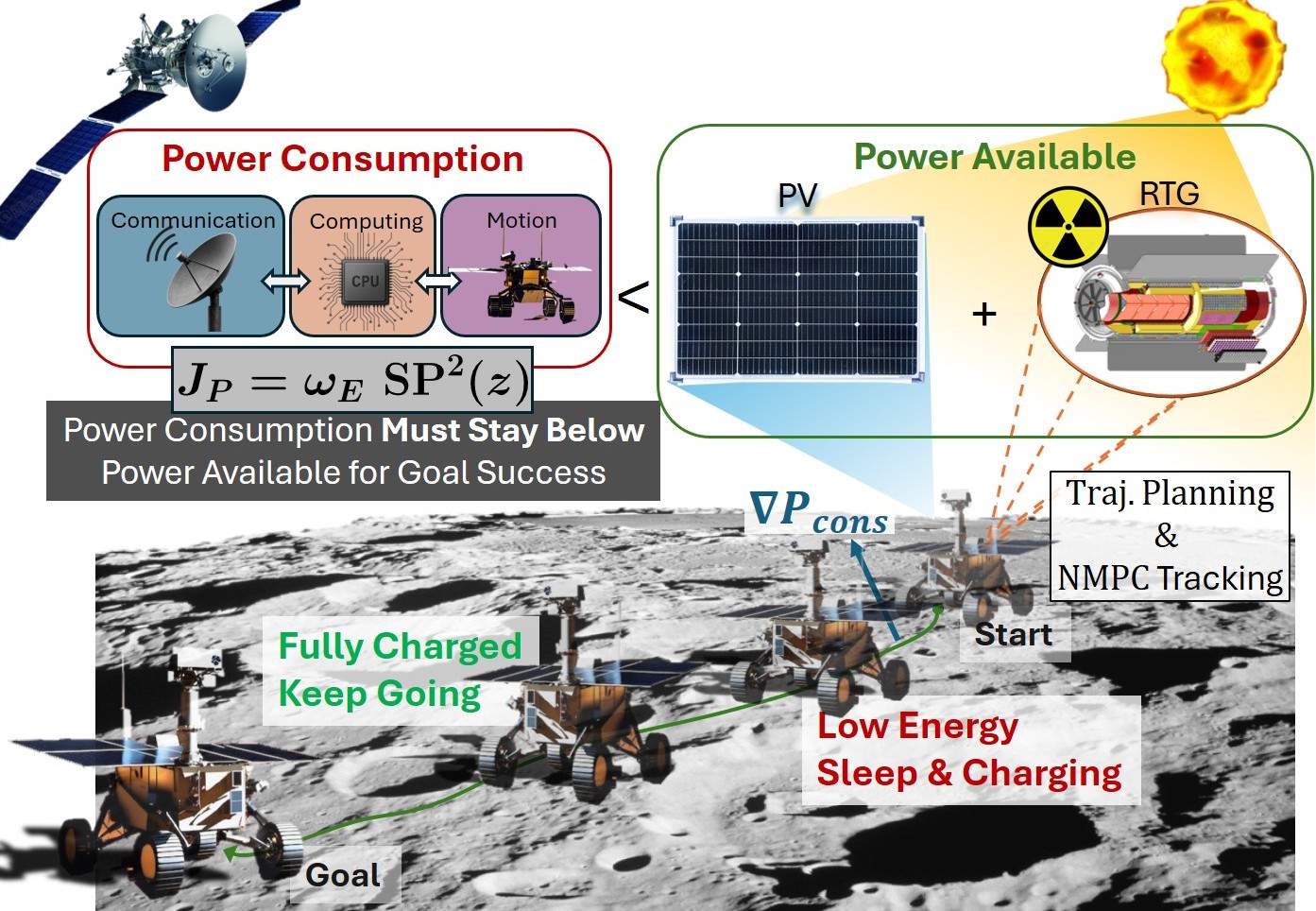}
    \caption{Overview of the energy-constrained planning framework ensuring power consumption stays within hybrid RTG–solar availability.}
    \label{fig:firstpage}
    \vspace{-15pt}
\end{figure}

Planetary exploration and long-range field robotics have recently gained significant momentum, driven by missions on the Moon, Mars, and extreme Earth environments~\cite{caraccio2025ardito, lanham2025planetary, li2023special, fonseca2021design}. These platforms must operate over extended durations with severely constrained energy resources, as shown in Fig. \ref{fig:firstpage}. In such scenarios, robots often rely on hybrid power sources—most notably, a low but steady output from radioisotope thermoelectric generators (RTGs)~\cite{clark2025comparison, tailin2024comprehensive, liu2023comprehensive}, combined with constant solar photovoltaic (PV) input~\cite{ndalloka2024solar, bamisile2025environmental, dada2023recent}. 
As the autonomy of these systems increases, so does the demand for intelligent decision-making that not only accounts for traversability \cite{yang2025fast} and safety, but also explicitly manages energy consumption. Despite its importance, energy-aware planning remains an under-addressed problem in the field.  

Existing works in robotic navigation and planning \cite{kiss2025sampling,kostyukov2025global,analooee2025dmtm,shang2025hgee,pan2025action,liu2025enhanced} largely neglect detailed energy modeling \cite{bouhabza2025energy}. Classical approaches, such as A*, D*, and their sampling-based variants, typically assume constant cost or incorporate only terrain-based metrics like slope or roughness~\cite{hedrick2020terrain, zhang2024review, zhang2025improved}.
Recent learning-based planners have made advances in perception and policy learning~\cite{xiao2022motion} but seldom consider power availability as a constraint. Some studies in long-range navigation or rover simulation incorporate energy-related costs~\cite{lamarre2020canadian}, but often treat them as external constraints rather than dynamically coupled factors. Moreover, perception and control modules—often significant energy consumers—are rarely integrated into the planning loop~\cite{mageshkumar2024adaptive, visca2021conv1d, malawade2022ecofusion, liu2025adaptive, li2025graph, yang2025imu, mao2025graph, du2022integrated, li2024ba, tang2025ba, li2024hcto, park2021elasticity, lv2023continuous, lang2023coco, talbot2025continuous, takemura2024perception}.
This disconnection can lead to suboptimal behavior in energy-constrained missions, where the robot may exhaust its power prematurely or avoid scientifically valuable regions due to conservative planning.  
In this work, we propose an optimization-based energy-constrained trajectory planning framework for planetary rovers operating under hybrid RTG and constant solar power inputs. Our method explicitly models both motion-related power consumption and the baseline subsystem loads, while representing available power as the combination of constant RTG output and fixed solar input. We formulate a trajectory optimization problem over SE(2) polynomial splines, incorporating both cumulative energy budgets and instantaneous power constraints. The planner also supports adaptive behaviors such as inserting waiting states to accumulate RTG energy when the available power is insufficient. Simulation results on Mars-like terrains demonstrate that the proposed framework generates smooth, feasible, and power-compliant trajectories, enabling robust autonomy in energy-limited environments.

The main contributions of this paper are as follows:
\begin{itemize}
    \item We present an energy-constrained trajectory planning framework for planetary rovers powered by hybrid RTG and constant solar input, explicitly modeling both motion-related power consumption and the baseline subsystem loads.  
    
    \item We develop physics-based models of power requirements for translational and rotational motion, as well as resistive effects on uneven planetary terrain, and incorporate them with baseline subsystem power within a unified trajectory planning formulation.
    
    \item We incorporate instantaneous power constraints into SE(2)-based polynomial trajectory optimization, enabling trajectories that satisfy power feasibility while preserving dynamic smoothness.
    
    \item We validate the proposed method in simulation, demonstrating power-feasible and dynamically smooth trajectories, and benchmark it against planners without power-limit enforcement. 

\end{itemize}

\section{Related Work}
\subsection{Energy-aware Navigation}
Energy-aware planning has been extensively studied for aerial and underwater vehicles, where trajectory generation is often formulated under battery constraints. For UAVs, minimum-energy paths and wind-aware models have been proposed~\cite{folk2025towards, taye2025enhancing, niaraki2025maximizing, kazemdehbashi2024exact}, while AUVs optimize drag-aware routes~\cite{zhang2024change}. For ground robots, energy cost is commonly approximated using elevation, slope, or terrain classes~\cite{nguyen20253d}, but power limits are seldom explicitly enforced~\cite{sakayori2021energy}. Some approaches include solar energy as a heuristic reward or penalty, but without modeling power flow or hard constraints on availability~\cite{plonski2016environment}. Overall, few works have directly incorporated global energy feasibility or instantaneous power limits into trajectory optimization.
\begin{figure}
    \centering
    \includegraphics[width=1\linewidth]{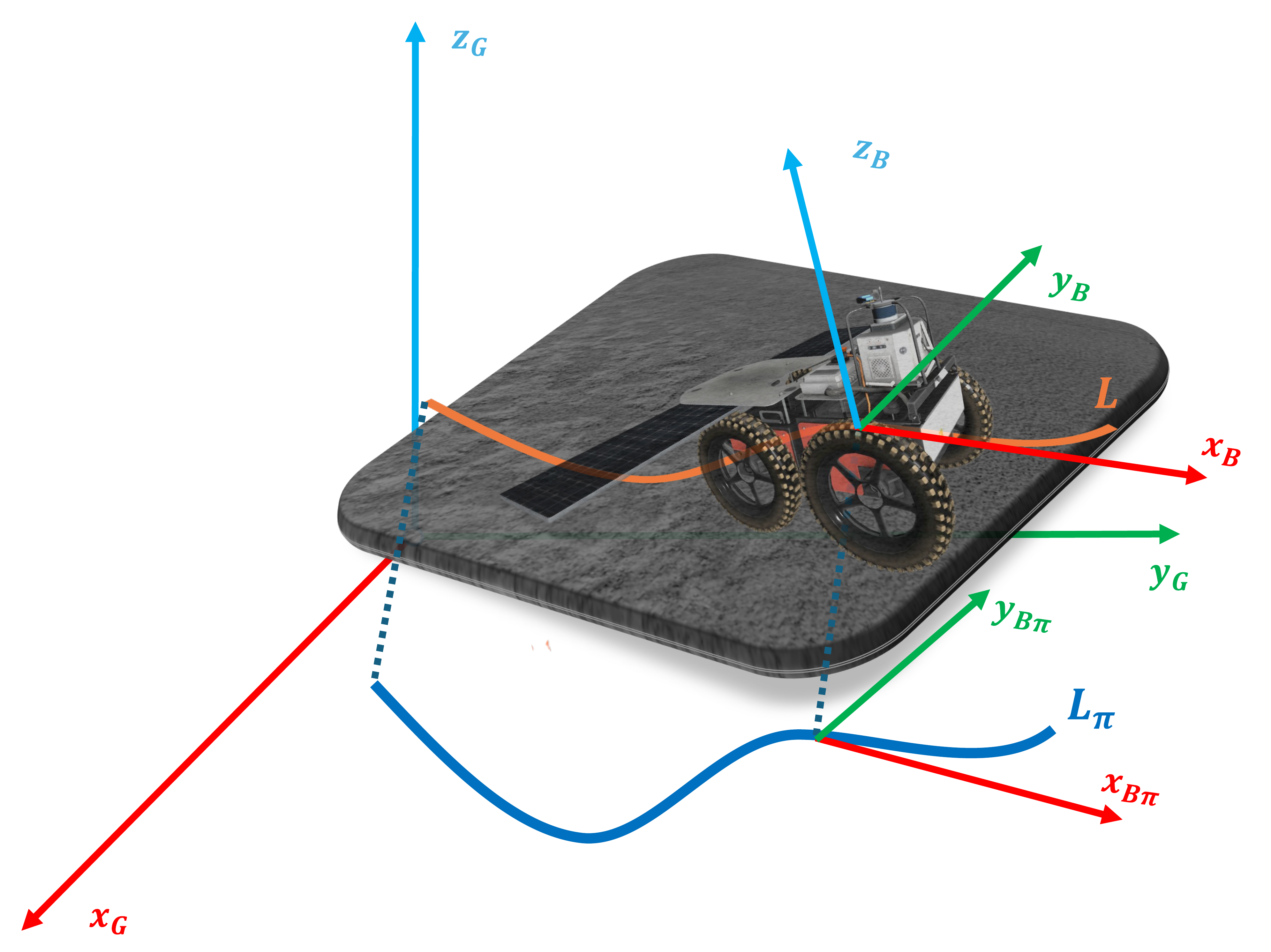}
    \caption{Diagram of Lunar Rover Coordinate System}
    \label{fig:Coordinates}
\end{figure}
\subsection{Terrain-Aware Planning}
Navigation in unstructured terrain typically relies on terrain-aware planning. Classical methods encode slope and roughness into cost maps~\cite{xu2023efficient, li2025real, weerakoon2023terrain, yoon2024analysis},
while more recent approaches employ learned traversability or semantic segmentation~\cite{kim2024learning, hosseinpoor2021traversability, mattamala2025wild, li2025seb}. However, terrain-induced energy penalties such as climbing resistance or inertia effects are rarely integrated into planning as explicit power constraints. In contrast, our work directly embeds the physics-based power model, including slope and motion dynamics, into the trajectory optimization process.

\subsection{Trajectory Optimization under Physical Constraints}
Trajectory optimization has been widely used in robotics to ensure feasibility under dynamic and environmental constraints. 
Polynomial splines, convex formulations, and sequential convex programming (SCP) have been employed to guarantee smoothness, collision avoidance, and bounded control inputs~\cite{ji2023convex, xu2021autonomous, scheffe2022sequential, zhao2024nurbs}. 
Model predictive control (MPC) frameworks are also commonly used to track optimized trajectories in real time, providing robustness to disturbances and model mismatch~\cite{song2023isolating, yang2023disturbance, gong2021lyapunov, mcallister2021inherent, li2025ua}. 

However, these approaches typically focus on kinematic and dynamic feasibility, emphasizing geometric accuracy or smooth control profiles, without explicitly considering power or energy constraints. In long-duration missions with severely limited onboard power, trajectories that are dynamically feasible but energetically infeasible may still lead to mission failure. In addition to motion-related power, a rover must continuously allocate baseline power to maintain sensing, computation, communication, and thermal management. The main \textbf{challenge} lies in jointly ensuring dynamic feasibility and energy feasibility under hybrid power inputs. Our work addresses this gap by integrating physics-based power models and hybrid energy constraints directly into the trajectory optimization process, while relying on MPC for accurate execution of the planned trajectory.



\section{Energy-Constrained Planning Framework}
This section introduces an energy-aware trajectory planning framework for planetary robots powered by hybrid Radioisotope Thermoelectric Generator (RTG) and solar photovoltaic (PV) energy systems. The framework integrates energy consumption models for motion, terrain resistance, and baseline subsystem power, ensuring trajectory feasibility under stringent energy constraints.

\subsection{Coordinate Frames and Kinematics}
\begin{figure}
    \centering
    \includegraphics[width=1\linewidth]{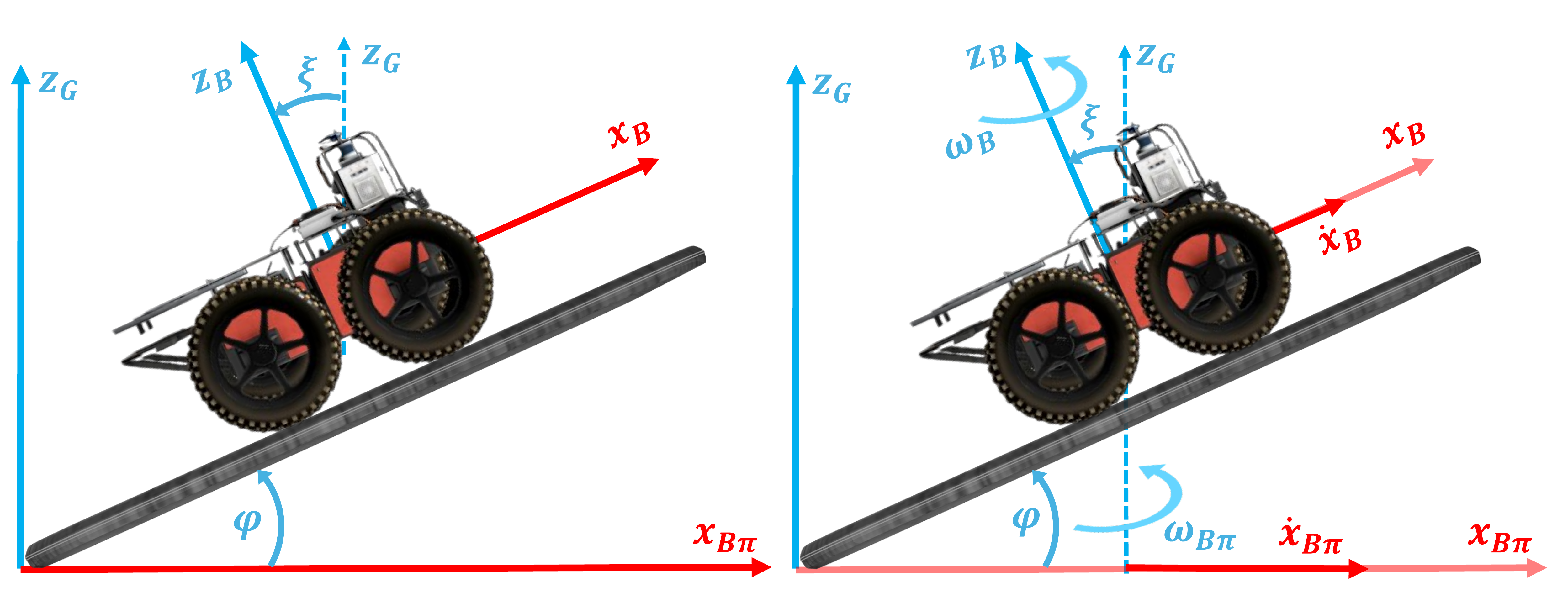}
    \caption{Diagram of Inclined Plane Coordinate System}
    \label{fig:InclinedCoordinates}
\end{figure}

We consider a differential-driven lunar rover navigating complex 3D terrain $\mathcal{M} \subset \mathbb{R}^3$, as shown in Fig.~\ref{fig:Coordinates}. Its motion is described in the global frame $\frameglobal=\{\globXvec,\globYvec,\globZvec\}$ and the body frame $\framebody=\{\vehicleXvec,\vehicleYvec,\vehicleZvec\}$. The projection of $\framebody$ onto the $\globXvec$–$\globYvec$ plane defines the 2D projected frame $\frameproj=\{\projXvec,\projYvec\}$.
The projection yaw angle $\globyaw$ is defined as the orientation of $\projXvec$ with respect to the global axis $\globXvec$.
Let $L_{\frG}$ denote the trajectory of the robot on the complex 3D terrain $\mathcal{M}$ under the global coordinate system $\frameglobal$, which is given by:
\[
L_{\frG} = \left\{ \mathbf{p}_{\frG}(t) \in \mathrm{SE}(3) \;\middle|\;
\begin{aligned}
\mathbf{p}_{\frG}(t) = (&\globX(t), \globY(t), \globZ(t), \\
                        &\globroll(t), \globpitch(t), \globyaw(t))
\end{aligned}
\right\},
\]

The projection of $L_{\frG}$ onto the 2D plane $\frameproj$ is given by:
\[
L_{\frP} = \left\{ \mathbf{p}_{\frP}(t) \in \mathrm{SE}(2) \;\middle|\; \mathbf{p}_{\frP}(t) = (\globX(t), \globY(t), \globyaw(t))\right\},
\]
where $t \in [0,T_f]$ denotes the planning horizon, with $T_f$ being the final time of the trajectory.

As shown in Fig.~\ref{fig:InclinedCoordinates}, 
$\varphi$ denotes the pitch angle, defined as the angle from $\projXvec$ to $\vehicleXvec$. 
Similarly, $\xi$ denotes the attitude angle, defined as the angle from $\globZvec$ to $\vehicleZvec$. 
The robot's velocity and acceleration along $\vehicleXvec$ are denoted by $\vehicleVX$ and $\vehicleAX$, 
and its angular velocity and acceleration around $\vehicleZvec$ are denoted by $\globW$ and $\globWd$, respectively. 
When projected onto $\frameproj$, these quantities are represented by $\globVX$, $\globVY$, $\globAX$, $\globAY$, $\globyaw$, $\globW$, and $\globWd$, respectively. 
Their relations to the original body-frame velocities are given by
\begin{equation}
    \vehicleVX = \frac{\sqrt{\globVX^2 + \globVY^2}}{\cos{\varphi}},
    \label{projv}
\end{equation}
\begin{equation}
    \vehicleAX = \frac{\globAX \cos{\globyaw} + \globAY \sin{\globyaw}}{\cos{\varphi}},
\end{equation}
\begin{equation}
    \vehicleW = \frac{\globW}{\cos{\xi}},
    \label{projw}
\end{equation}
\begin{equation}
    \vehicleWd = \frac{\globWd}{\cos{\xi}}.
\end{equation}


\subsection{Problem Statement}
The rover is powered by a hybrid energy source consisting of constant RTG power and solar PV input. 
The instantaneous power consumption $P_{\text{cons}}$ is modeled as:
\begin{equation}
P_{\text{cons}} = P_{\text{mot}} + P_{\text{base}},
\end{equation}
where $P_{\text{mot}}$ denotes the motion power consumption, and $P_{\text{base}}$ represents the base power required to maintain baseline onboard systems. 
The total available power at time $t$ is given by:
\begin{equation}
P_{\text{avail}}(t) = P_{\text{RTG}}(t) + P_{\text{solar}}(t).
\end{equation}
The trajectory must satisfy the following feasibility constraints:
\begin{equation}
P_{\text{cons}}(t) \leq P_{\text{avail}}(t), 
\quad \forall t \in [0,T_f].
\end{equation}



\subsection{Motion Power Modeling}
The motion power $P_{\text{mot}}(t)$ consists of three components, the linear motion power $P_{\text{lin}}(t)$, the rotational motion power $P_{\text{rot}}(t)$, and the resistance power $P_{\text{res}}(t)$:
\begin{equation}
    P_{\text{mot}}(t) = P_{\text{lin}}(t) + P_{\text{rot}}(t) + P_{\text{res}}(t).
\end{equation}

\paragraph{Linear motion power} 
This term depends on the robot mass $m$, the longitudinal velocity $\vehicleVX(t)$, and the longitudinal acceleration $\vehicleAX(t)$. It is expressed as:
\begin{equation}
    P_{\text{lin}}(t) = m \bigl(\vehicleAX(t) + g \sin{\varphi(t)}\bigr)\,\vehicleVX(t).
\end{equation}

\paragraph{Rotational motion power} 
This term depends on the moment of inertia $I_z$ about the body $\vehicleZvec$ axis, the angular velocity $\vehicleW(t)$, and the angular acceleration $\vehicleWd(t)$. It is expressed as:
\begin{equation}
    P_{\text{rot}}(t) = I_z \,\vehicleWd(t)\,\vehicleW(t).
\end{equation}

\paragraph{Persistent resistive power}
In addition to inertial and gravitational terms, the rover experiences persistent resistive forces that increase with velocity, including rolling resistance, viscous damping, and terrain-induced drag~\cite{kimura2007adaptive, wong2022theory}.
The corresponding power consumption is modeled as:
\begin{equation}
    P_{\text{res}}(t) = \bigl( C_0 + C_1 \, |\vehicleVX(t)| + C_2 \, \vehicleVX^2(t) \bigr)\, \vehicleVX(t),
\end{equation}
where $C_0$, $C_1$, and $C_2$ are empirical coefficients determined by terrain properties. On the lunar surface, aerodynamic drag is negligible
($\rho \approx 0$), so $C_2$ mainly reflects additional soil deformation
losses. This term represents an ever-present component of energy dissipation that increases nonlinearly with forward velocity.

\paragraph{Trajectory Projection onto the 2D Plane} 
Since the planned trajectory $L_{\frP}$ lies in the projected 2D plane, 
the expressions of $P_{\text{lin}}$ and $P_{\text{rot}}$ must be reformulated accordingly. 
Using the relationships between projected $\frameproj$ and global $\frameglobal$ frame velocities and accelerations, we first introduce some shorthand. For compactness, define
\begin{equation}
\begin{aligned}
a &= \globAX \cos{\globyaw} + \globAY \sin{\globyaw}, \\[4pt]
b &= -\globAX \sin{\globyaw} + \globAY \cos{\globyaw}, \\[4pt]
v_{\frG} &= \sqrt{\globVX^2 + \globVY^2}, \\[4pt]
s &= \sin{\varphi}, \qquad
c = \cos{\varphi}.
\end{aligned}
\end{equation}

With these, the linear and rotational motion power are
\begin{equation}
    P_{\text{lin}} = m \left( \frac{a}{\cos{\varphi}} + g \sin{\varphi} \right)\frac{v_{\frG}}{\cos{\varphi}},
\end{equation}
\begin{equation}
    P_{\text{rot}} = I_z \,\frac{\globWd\,\globW}{\cos{\xi}}.
\end{equation}

The gradients of $P_{\text{lin}}$ with respect to global states and $\varphi$ are:
\begin{equation}
\nabla P_{\text{lin}} =
\begin{bmatrix}
\dfrac{\partial P_{\text{lin}}}{\partial \globVX} \\[10pt]
\dfrac{\partial P_{\text{lin}}}{\partial \globVY} \\[10pt]
\dfrac{\partial P_{\text{lin}}}{\partial \globAX} \\[10pt]
\dfrac{\partial P_{\text{lin}}}{\partial \globAY} \\[10pt]
\dfrac{\partial P_{\text{lin}}}{\partial \globyaw} \\[10pt]
\dfrac{\partial P_{\text{lin}}}{\partial \varphi}
\end{bmatrix}
=
\begin{bmatrix}
m\!\left(\tfrac{a}{c}+g s\right)\dfrac{\globVX}{c \,v_{\frG}} \\[10pt]
m\!\left(\tfrac{a}{c}+g s\right)\dfrac{\globVY}{c \, v_{\frG}} \\[10pt]
m\,\dfrac{v_{\frG}}{c^{2}}\,\cos{\globyaw} \\[10pt]
m\,\dfrac{v_{\frG}}{c^{2}}\,\sin{\globyaw} \\[10pt]
m\,\dfrac{v_{\frG}}{c^{2}}\,b \\[10pt]
m\,v_{\frG}\!\left(\dfrac{2as}{c^{3}} + \dfrac{g}{c^{2}}\right)
\end{bmatrix}.
\end{equation}

Similarly, the gradients of $P_{\text{rot}}$ with respect to $\globW, \globWd$, and $\xi$ are:
\begin{equation}
\nabla P_{\text{rot}} =
\begin{bmatrix}
\dfrac{\partial P_{\text{rot}}}{\partial \globW} \\[8pt]
\dfrac{\partial P_{\text{rot}}}{\partial \globWd} \\[8pt]
\dfrac{\partial P_{\text{rot}}}{\partial \xi}
\end{bmatrix}
=
\begin{bmatrix}
I_z \dfrac{\globWd}{\cos{\xi}} \\[10pt]
I_z \dfrac{\globW}{\cos{\xi}} \\[10pt]
I_z \dfrac{\globWd \,\globW \,\sin{\xi}}{\cos^2{\xi}}
\end{bmatrix}.
\end{equation}

\subsection{Baseline Power Modeling}
In addition to motion-related power, the rover requires the baseline power consumption to sustain sensing, computation, communication, and thermal management. 
These components are aggregated into a constant term:
\begin{equation}
P_{\text{base}} = P_{\text{perc}} + P_{\text{plan}} + P_{\text{ctrl}} + P_{\text{misc}},
\end{equation}
where $P_{\text{perc}}$, $P_{\text{plan}}$, $P_{\text{ctrl}}$, and $P_{\text{misc}}$ denote the average steady-state power draws of the perception, planning, control, and miscellaneous subsystems, respectively. 
Since these loads are approximately constant during operation, $P_{\text{base}}$ is treated as a fixed parameter in the subsequent modeling and optimization.

\subsection{Smooth Power-Limit Penalty and Gradient Derivation}
We penalize violations of the instantaneous power constraint 
$P_{\text{cons}}(t) \leq P_{\text{avail}}(t)$ 
using a smooth hinge (softplus) formulation~\cite{bishop1995neural, dugas2000incorporating}. 
Directly enforcing this inequality in gradient-based optimization may lead to discontinuities or abrupt changes in the gradient. 
The softplus function provides a continuously differentiable approximation to the ReLU hinge, 
thereby mitigating gradient discontinuities and improving optimization stability.
Let
\begin{equation}
    S(t) = \sqrt{P_{\text{lin}}^2(t) + P_{\text{rot}}^2(t)},
\end{equation}
\begin{equation}
    \tau = P_{\text{avail}} - P_{\text{cons}},
\end{equation}
The softplus function $\text{SP}(\cdot)$ and its associated logistic sigmoid $\sigma(\cdot)$ are defined as
\begin{equation}
\begin{cases}
\text{SP}(z) = \dfrac{1}{\kappa}\ln\!\bigl(1+e^{\kappa z}\bigr), \\[6pt]
\sigma(z)   = \dfrac{1}{1+e^{-z}}, \\[6pt]
z = S(t) - \tau.
\end{cases}
\end{equation}
The penalty cost is
\begin{equation}
J_P = \omega_E \,\text{SP}^2(z),
\end{equation}
where $\omega_E > 0$ is the penalty weight. 

The key gradients are
\begin{equation}
\frac{dJ_P}{dz} = 2\,\omega_E\,\text{SP}(z)\,\sigma(\kappa z).
\end{equation}
Since $S(t) = \sqrt{P_{\text{lin}}^2(t) + P_{\text{rot}}^2(t)}$, we have
\begin{equation}
\frac{\partial J_P}{\partial P_{\text{lin}}}
= \frac{dJ_P}{dz}\,\frac{P_{\text{lin}}}{S}, 
\end{equation}
\begin{equation}
\frac{\partial J_P}{\partial P_{\text{rot}}}
= \frac{dJ_P}{dz}\,\frac{P_{\text{rot}}}{S}.
\end{equation}

Applying the chain rule to the physical variables gives
\begin{equation}
\nabla J_P
=
\frac{\partial J_P}{\partial P_{\text{lin}}}\,\nabla P_{\text{lin}}
\;+\;
\frac{\partial J_P}{\partial P_{\text{rot}}}\,\nabla P_{\text{rot}}.
\end{equation}

It is worth noting that 
$\dfrac{\partial P_{\text{lin}}}{\partial \varphi}$ 
and 
$\dfrac{\partial P_{\text{rot}}}{\partial \xi}$ 
can be further transformed into gradients with respect to the trajectory $L_{\frP}$. 
We omit the detailed derivation here for brevity. Thus, the gradient of the power-penalty cost $J_P$ with respect to all physical variables can be obtained directly from the previously derived expressions of $\nabla P_{\text{lin}}$ and $\nabla P_{\text{rot}}$.
\subsection{Trajectory Optimization Under Energy Constraints}
\label{subsec:energy_opt}
The problem is to find a trajectory $L_{\frP}$, obtained by projecting $L_{\frG}$ onto the plane $\frP$, where $L_{\frG}$ is dynamically feasible on uneven terrain and satisfies the hybrid power budget. The trajectory is parameterized by quintic polynomial splines with coefficients $c$ and segment durations $T_i$, with the total horizon $T_f=\sum_i T_i$.
\paragraph{Objective}
The cost functional combines smoothness, temporal regularization, terrain difficulty, and energy feasibility:
\begin{equation}
\begin{aligned}
\min_{c,T_i}\quad
&\int_0^{T_f}\!\!\|j(t)\|^2\,dt
+\rho_T T_i
+\rho_r\,\mathcal{R}(L_{\frG})
+J_P,  \\
\text{s.t. }\quad
& \tau = P_{\text{avail}} - P_{\text{cons}}, \;  T_i>0, \\
& \mathbf{p}(0)=\mathbf{p}_0,\; \mathbf{p}(T_f)=\mathbf{p}_g, \\
& \vehicleVX \le \dot{x}_{\frB \max}, \;
|\vehicleAX| \le \ddot{x}_{\frB \max}, \;
|\vehicleAY| \le \ddot{y}_{\frB \max},
\end{aligned}
\label{eq:energy_obj}
\end{equation}
where $j(t)$ denotes the trajectory jerk, $\mathcal{R}$ represents the accumulated terrain risk~\cite{xu2023efficient}, $T_i$ penalizes the overall trajectory duration, and $J_P$ is the softplus-based penalty enforcing instantaneous power feasibility. Here, $\mathbf{p}_0$ and $\mathbf{p}_g$ denote the initial and goal states, respectively.
\begin{figure*}
    \centering
    \includegraphics[width=1\linewidth]{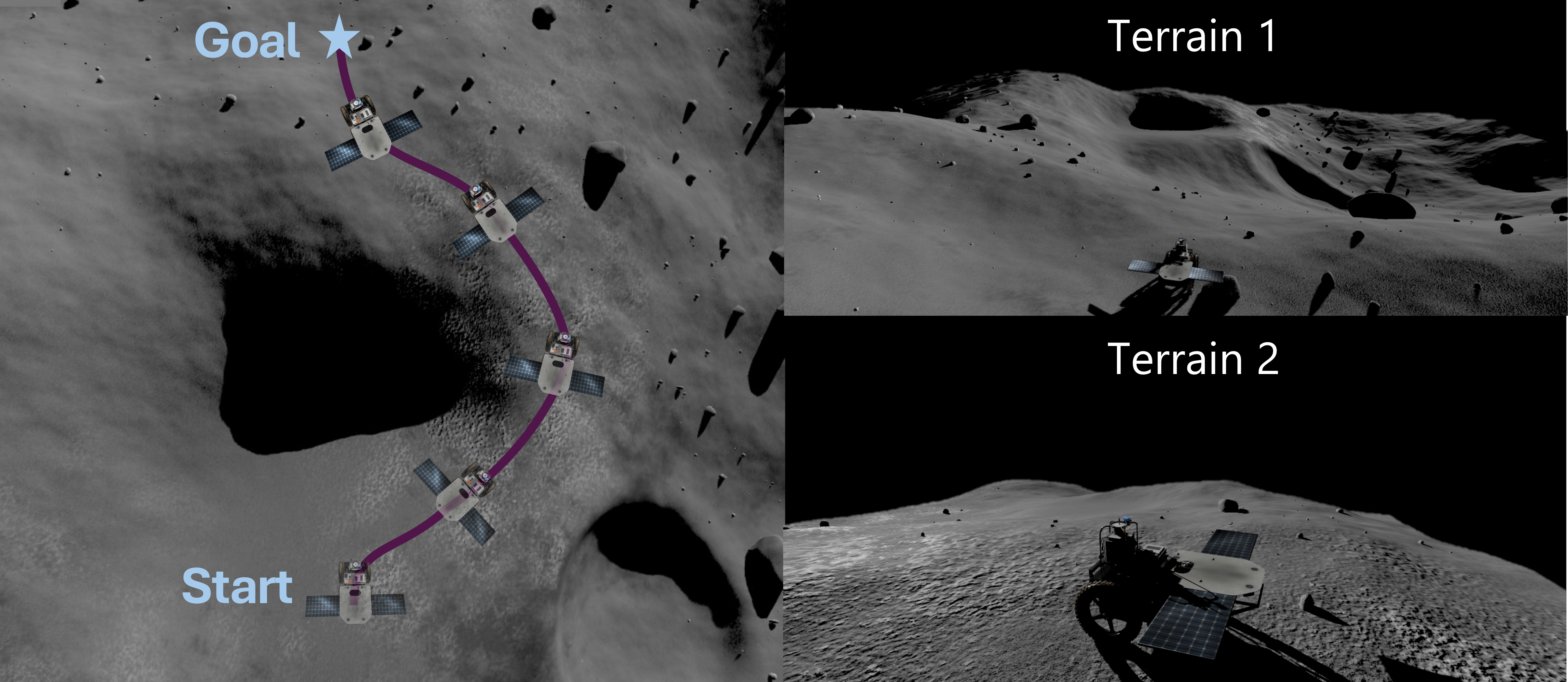}
    \caption{The lunar rover tracks the planned trajectory while respecting the power limit.}
    \label{fig:simulation}
\end{figure*}
\paragraph{Energy budget}
In addition to the soft penalty $J_E$, a global budget constraint may be imposed:
\begin{equation}
\int_0^{T_f} P_{\mathrm{cons}}(t)\,dt \;\le\; \int_0^{T_f} P_{\mathrm{avail}}(t)\,dt.
\end{equation}
If no continuous feasible trajectory exists, the planner may insert zero-velocity spline 
segments as passive waiting intervals to accumulate RTG energy without advancing spatially.

\paragraph{Solution scheme}
The problem \eqref{eq:energy_obj} is solved via sequential convex programming (SCP) with augmented Lagrangian handling of inequality constraints, ensuring that the resulting trajectories are dynamically feasible, terrain-aware, and compliant with the hybrid energy budget.

\section{NMPC Tracking for Differential-Driven Robot}
To enable accurate execution of the planned trajectory $L_{\frP}$ on complex terrain, 
we employ a nonlinear model predictive control (NMPC) strategy. 
At each control cycle, the reference spline trajectory $L_{\frP}$ is discretized into a prediction horizon of $N_p$ steps, 
while the control horizon is limited to $N_c$ steps, after which the control inputs are held constant.

We extract the reference positions $(\globX,\globY,\globyaw)$ at each discretized prediction step. The state vector is defined as:
\begin{equation}
X_{\frP}(k) = 
\begin{bmatrix}
\globX(k), \; \globY(k), \; \globyaw(k)
\end{bmatrix}^\top,
\end{equation}  
and the resulting control vector in the projection frame $\frP$ is given by:
\begin{equation}
u_{\frP}(k) =
\begin{bmatrix}
\projVX(k), \; \projW(k)
\end{bmatrix}^\top.
\end{equation}

To ensure that the controller follows not only the geometric path but also the expected time-varying power demand of the reference trajectory, we enforce tracking of the reference velocities.
Specifically, from $L_{\frP}$ we obtain the projected velocity components $(\globVX,\globVY,\globW)$. 
These are combined into the projected control reference:
\begin{equation}
u^{\mathrm{ref}}_{\frP}(k) =
\begin{bmatrix}
\projVX^{\mathrm{ref}}(k) \\[4pt] \projW^{\mathrm{ref}}(k)
\end{bmatrix}
=
\begin{bmatrix}
\sqrt{\globVX^2(k)+\globVY^2(k)} \\[4pt]
\globW(k)
\end{bmatrix},
\end{equation}
which is subsequently used in the cost function to penalize velocity tracking errors.

The predictive controller is constructed as a finite-horizon optimization problem:  
\begin{equation}
\min_{U} \;
\| X_{\frP} - X_{\frP}^{\mathrm{ref}} \|_{Q}^2
+ \| U_{\frP} - U_{\frP}^{\mathrm{ref}} \|_{R}^2
+ \| \Delta U_{\frP} \|_{R_d}^2,
\label{eq:nmpc_cost_compact}
\end{equation}
where $Q,R,R_d$ are positive-definite weighting matrices for state tracking, 
velocity tracking, and control rate variation, respectively. 
The stacked vectors are defined as:
\begin{equation}
\begin{aligned}
X_{\frP} &= \begin{bmatrix} X_{\frP}^\top(0) & X_{\frP}^\top(1) & \cdots & X_{\frP}^\top(N_p) \end{bmatrix}^\top, \\[6pt]
U_{\frP} &= \begin{bmatrix} u_{\frP}^\top(0) & u_{\frP}^\top(1) & \cdots & u_{\frP}^\top(N_c) \end{bmatrix}^\top, \\[6pt]
\Delta U_{\frP} &= \begin{bmatrix}
\Delta u_{\frP}^\top(1) & \Delta u_{\frP}^\top(2) & \cdots & \Delta u_{\frP}^\top(N_c)
\end{bmatrix}^\top,
\end{aligned}
\end{equation}
where the reference trajectories $X_{\frP}^{\mathrm{ref}}$ and $U_{\frP}^{\mathrm{ref}}$ 
are constructed analogously from $L_{\frP}$, 
and the control increment is defined as 
$\Delta u_{\frP}(k) = u_{\frP}(k) - u_{\frP}(k-1)$ at step $k$.

The optimization problem is implemented using the CasADi framework, which 
provides automatic differentiation and interfaces with multiple nonlinear solvers. 
At each control cycle, the NMPC yields the optimal control sequence, from which only the first element is applied:  
\[
u_{\frP}^*(k) = 
\begin{bmatrix} \projVX(k) \\ \projW(k) \end{bmatrix}.
\]
Since this solution is obtained in the projection frame $\frP$, it must be mapped into the body frame $\frB$ for execution. 
Using the projection relations defined in Eq.~\eqref{projv} and \eqref{projw}, 
the body-frame control command is:
\[
u_{\frB}^*(k) = 
\begin{bmatrix} \vehicleVX(k) \\ \vehicleW(k) \end{bmatrix}
=
\begin{bmatrix}
\dfrac{\projVX(k)}{\cos\varphi(k)} \\[8pt]
\dfrac{\projW(k)}{\cos\xi(k)}
\end{bmatrix}.
\]
This transformation ensures that the computed input respects the vehicle’s true longitudinal velocity and angular rate on uneven terrain, 
thereby closing the control loop.
\begin{figure}
    \centering
    \includegraphics[width=1\linewidth]{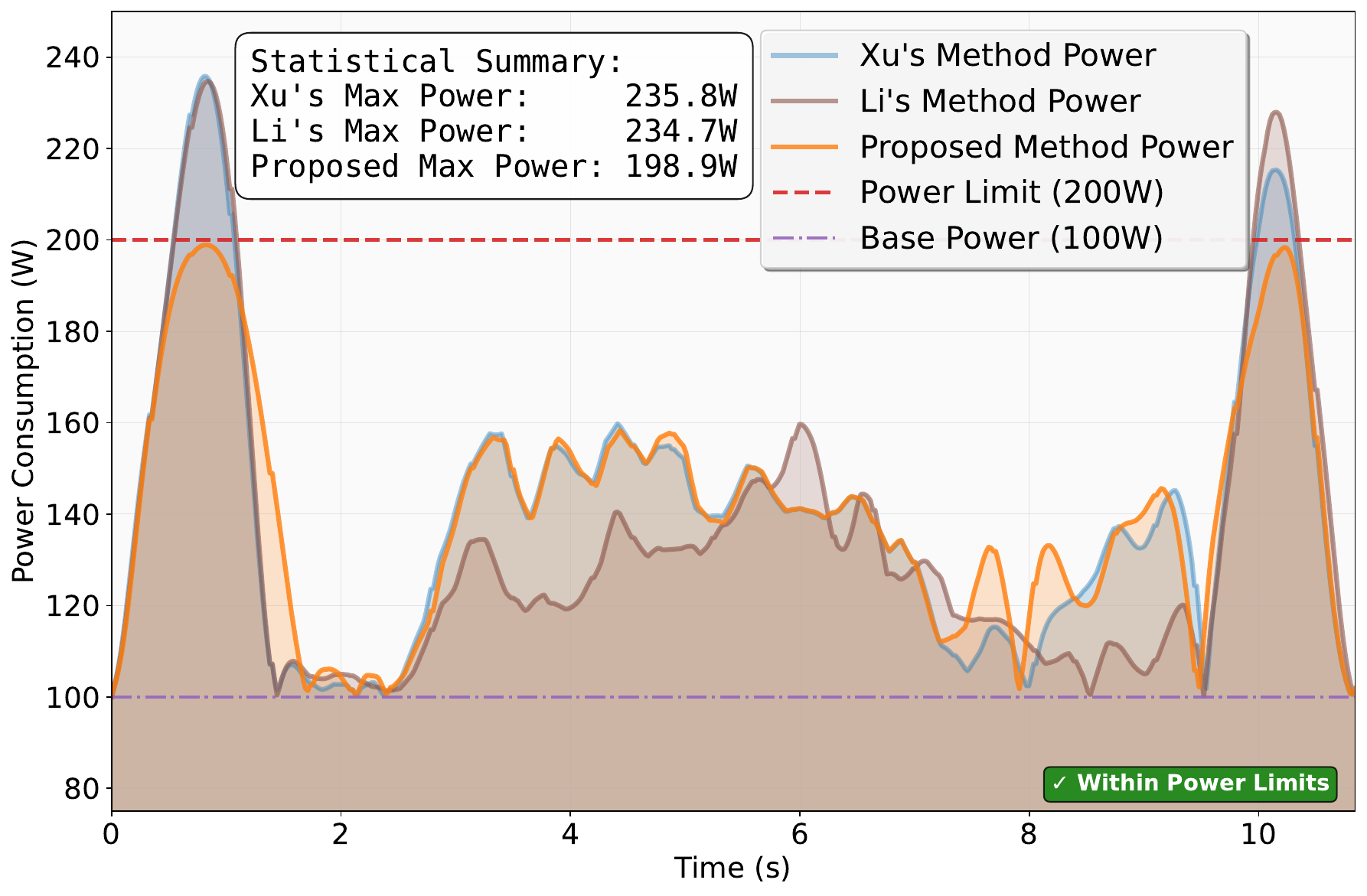}
    \caption{Planning Stage Power Comparison with Baseline}
    \label{fig:power_planning}
\end{figure}
\begin{table}[t]
\centering
\renewcommand{\arraystretch}{1.3} 
\caption{Comparison of Metrics for Different Methods}
\begin{tabular}{lccccc}
\hline
\hline
\textbf{Method} & \textbf{$P_{\text{pmax}}$ ($W$)} & \textbf{$P_{\text{tmax}}$ ($W$)} & \textbf{$e_y$ ($m$)} & \textbf{$e_{vx}$ ($m/s$)} \\ \hline 
 Xu et al.'s\cite{xu2023efficient} & 
 235.8 &
 268.5 &
 0.09 &
 0.10\\ \hline
 Li et al.'s \cite{li2025seb} & 
 234.7 &
 287.3 &
 \textbf{0.08} &
 0.13\\ \hline
Proposed &
\textbf{198.9} &
\textbf{202.52} &
 0.10 &
\textbf{0.05}\\ \hline
\end{tabular}
\label{tab:comparison}
\vspace{-15pt}
\end{table}
\section{Simulation}
\subsection{System Setup}
This work focuses on a differential-driven lunar rover robot. 
Due to hardware limitations, the evaluation is conducted entirely in simulation. 
As illustrated in Fig.~\ref{fig:simulation}, a lunar terrain with rocks, craters, and uneven surfaces is constructed in Gazebo-Sim-9 using ROS-Rolling (Ubuntu 24.04).
The simulated rover is differential-driven, with an overall length of approximately $1.2\,\mathrm{m}$ 
and a width of $2\,\mathrm{m}$ including solar panels, and has a total mass of $150\,\mathrm{kg}$. 
The objective is to generate a trajectory that respects an instantaneous power limit of $200\,\mathrm{W}$, 
which is subsequently tracked using a nonlinear model predictive control (NMPC) controller. 
All simulations are performed on a desktop equipped with a 13th Gen Intel Core i7 processor.
\begin{figure}
    \centering
    \includegraphics[width=1\linewidth]{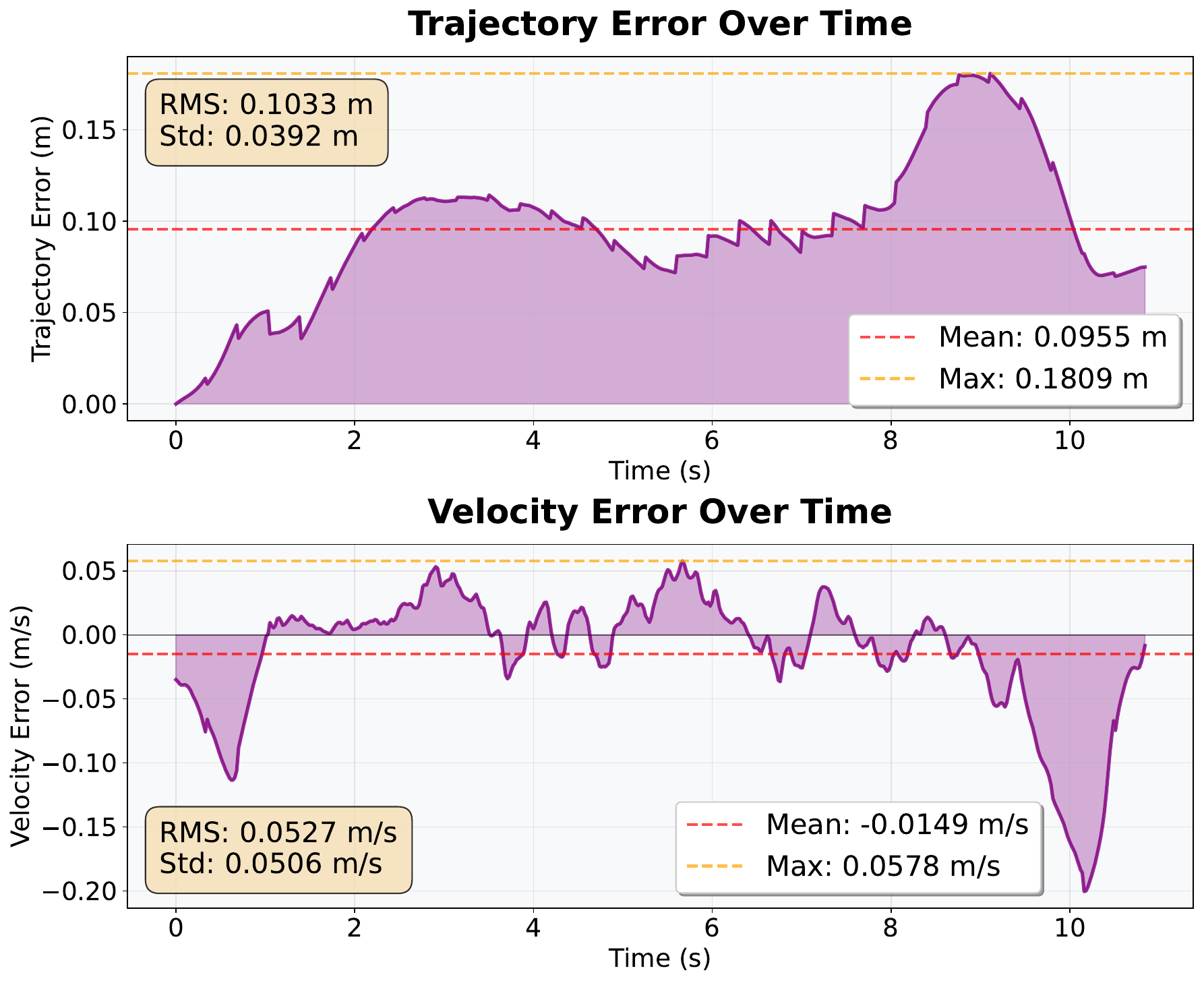}
    \caption{Tracking position error $e_y$ and velocity error $e_{vx}$ between the planned trajectory and the actual rover path during MPC tracking}
\label{fig:lateral_error}
\vspace{-10pt}
\end{figure}
\begin{figure}
    \centering
    \includegraphics[width=1\linewidth]{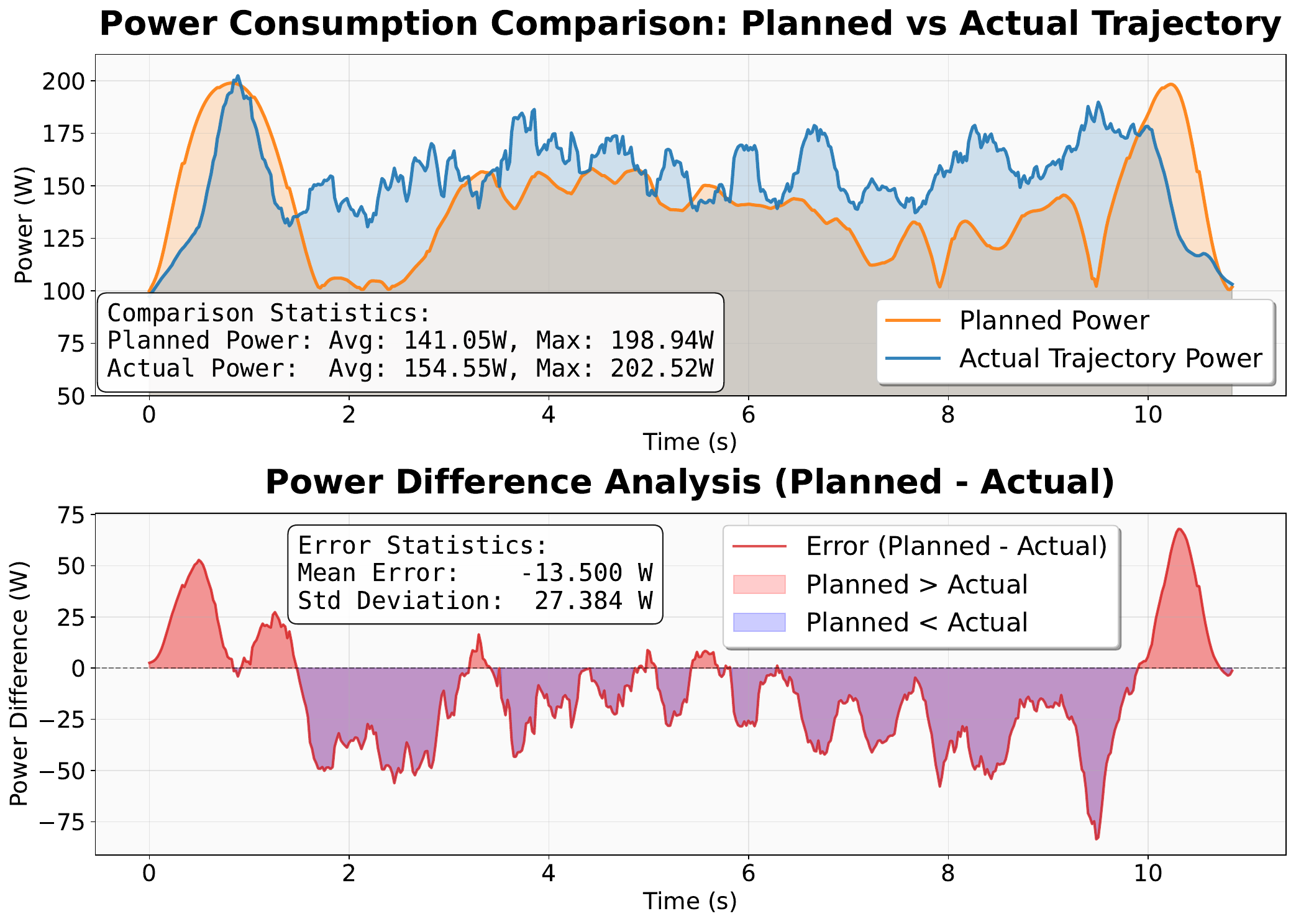}
    \caption{Comparison of Planned and Actual Power Consumption}
    \label{fig:power_compare}
    \vspace{-15pt}
\end{figure}
\subsection{Evaluation Metric}
The evaluation metric for our work is listed as follows:
\begin{itemize}
    \item \textbf{Maximum Power in Planning $P_{\text{pmax}}$:} 
    the peak power predicted along the planned trajectory.

    \item \textbf{Maximum Power in Tracking $P_{\text{tmax}}$:} 
    the highest instantaneous power consumption measured during NMPC tracking. 

    \item \textbf{Tracking Position Error (RMS) $e_y$:} 
    lateral position error $e_y$ between the planned trajectory and the actual rover path. 

    \item \textbf{Tracking Velocity Error (RMS) $e_{vx}$:} 
    the deviation of the rover’s forward velocity from the reference velocity along the trajectory.
\end{itemize}

\subsection{Result and Discussion}
Figure~\ref{fig:power_planning} and Table~\ref{tab:comparison} compare the proposed planning method with Xu et al.'s~\cite{xu2023efficient} and Li et al.'s~\cite{li2025seb}, both of which do not enforce a maximum power limit. The rover’s base power is specified as 100 W, with a planning constraint that caps the total power at 200 W.
Our method produces a trajectory with a peak power of 198.9 W (within the limit; 1.1 W margin, 0.55\%), whereas Xu et al.’s reaches 235.8 W (35.8 W over the cap; 17.9\%), and Li et al.’s reaches 234.7 W (34.7W over the cap; 17.4\%). This demonstrates that our method uniquely enforces power feasibility, while existing planners generate trajectories that violate energy limits.

In Fig.~\ref{fig:lateral_error}, the proposed tracking method is evaluated against the planned trajectory. It achieves a root-mean-square (RMS) lateral position error \(e_y\) of 0.1033 m and a (RMS) velocity error \(e_{vx}\) of 0.05 m/s. Because our method explicitly considers the reference velocity, it yields the smallest \(e_{vx}\). However, for the same reason, the lateral error \(e_y\) is slightly larger than Xu’s (0.09 m) and Li’s (0.08 m), though the differences remain acceptable. Overall, this highlights a trade-off: explicitly tracking reference velocities ensures power feasibility but introduces slightly higher lateral deviations.

Fig.~\ref{fig:power_compare} compares the planned power profile with the actual power, estimated from IMU and other sensor data during MPC tracking. Although the overall trends are consistent, discrepancies occur due to the difficulty of accurately modeling lunar resistance and the challenge of strictly following the planned acceleration profile. The peak actual power reaches 202.52~W, exceeding the 200~W limit by 2.52~W (1.26\%), indicating a minor violation.

\section{Conclusion}
This paper presents an energy-constrained trajectory planning framework for long-duration mobile robots operating under limited hybrid RTG and solar power. 
By explicitly modeling motion-related power consumption together with a baseline subsystem load and incorporating both cumulative and instantaneous power constraints, 
our approach enables robust and feasible navigation in power-limited environments. 
The framework builds on SE(2)-based trajectory optimization to jointly consider terrain dynamics and overall energy feasibility. Experimental results validate the effectiveness of the proposed method in generating smooth and power-compliant trajectories across varying terrain conditions. Future work includes real-world deployment on embedded platforms and integration with higher-level mission planners. The code and simulation platform will be released as open source upon publication.

\bibliographystyle{IEEEtran}
\bibliography{mybib}


\end{document}